\documentclass[journal]{IEEEtran}
\usepackage{graphicx}
\ifCLASSINFOpdf
\else
\fi
\usepackage{array}
\usepackage{mdwmath}
\usepackage{mdwtab}
\usepackage{eqparbox}
\usepackage{amsmath}
\usepackage{graphicx}
\usepackage{subfigure} 
\usepackage{booktabs}
\usepackage{threeparttable}
\usepackage{amsmath}
\usepackage{subfigure}
\DeclareMathOperator*{\argmax}{argmax}

\begin{document}
\title{An Efficient Probabilistic Solution to Mapping Errors in LiDAR-Camera Fusion for Autonomous Vehicles}
%
%
%
\author{Dan Shen,
        Zhengming Zhang,
        Renran Tian,
        Yaobin Chen,
        Rini Sherony
\thanks{D. Shen and Y. Chen are with the Department
of Electrical and Computer Engineering, Indiana University Purdue University Indianapolis, Indianapolis,
IN, 46202 USA (E-mails: {danshen,ychen}@iupui.edu)}
\thanks{Z. Zhang is with the School of Industrial Engineering, Purdue University, West Lafayette, IN, 47906 USA (E-mail: zhan3988@purdue.edu)}
\thanks{R. Tian is with the Department of Computer Information \& Graphics Technology, Indiana University Purdue University Indianapolis, Indianapolis,
IN, 46202 USA (E-mail: rtian@iupui.edu)}
\thanks{R. Sherony is with the Collaborative Safety Research Center, Toyota Motor Engineering and Manufacturing North America Inc., Ann Arbor, Michigan, USA (E-mail: rini.sherony@toyota.com)}
}

%
%

\maketitle

\markboth{}
{Shen \MakeLowercase{\textit{et al.}}: An Efficient Probabilistic Solution to Mapping Errors in LiDAR-Camera Fusion for Autonomous Vehicles}
%




\begin{abstract}
LiDAR-camera fusion is one of the core processes for the perception system of current automated driving systems. The typical sensor fusion process includes a list of coordinate transformation operations following system calibration. Although a significant amount of research has been done to improve the fusion accuracy, there are still inherent data mapping errors in practice related to system synchronization offsets, vehicle vibrations, the small size of the target, and fast relative moving speeds. Moreover, more and more complicated algorithms to improve fusion accuracy can overwhelm the onboard computational resources, limiting the actual implementation. This study proposes a novel and low-cost probabilistic LiDAR-Camera fusion method to alleviate these inherent mapping errors in scene reconstruction. By calculating shape similarity using KL-divergence and applying RANSAC-regression-based trajectory smoother, the effects of LiDAR-camera mapping errors are minimized in object localization and distance estimation. Designed experiments are conducted to prove the robustness and effectiveness of the proposed strategy.
\end{abstract}


%
\IEEEpeerreviewmaketitle

\section{Introduction}
%
%
%
%
\IEEEPARstart{A}UTONOMOUS vehicles (AVs) are more popular and attractive with the tremendous and impressive technological progress in recent years. Noticeable achievements in both software and hardware have brought more automation functionality into reality. As increasingly believe, AVs are promising to avoid potential crashes and protect human lives. In addition to driving safety, AVs are also expected to improve ride comfort and smoothen the traffic flow with an increasing level of service on the roads \cite{8}. According to the report of the U.S. Department of Transportation, around $94\%$ of accidents take place due to human errors, and over 90$\%$ of those accidents were caused by visual information acquisition problems \cite{4}\cite{5}. When a large percentage of severe and fatal crashes are related to vulnerable road users,  research topics in vulnerable road users (VRU) safety are an indispensable part of the advent of AVs driving in cities. It is vital to investigate and examine whether AVs will detect and be beneficial to the VRUs\cite{6}\cite{7}\cite{26}.

However, we are still far away from fully autonomous driving due to technical difficulties and legal ethics. Automated driving in mixed-traffic environments is one of the main barriers for the development of fully AVs, especially for the safety issues in the interactions among AVs, human-driven vehicles, and VRUs such as cyclists and pedestrians \cite{9}. Recently, the safety issues of micro-mobility platforms have also drawn increasing attention to the public. Accompanying the convenience of the trips, there have been numerous injuries and fatalities reported in association with micromobility options like e-scooters,  according to the estimations from the Centers for Disease Control and Prevention (CDC) \cite{10}. 

Although different companies or research groups may have different AV systems, most solutions generally have three main parts: perception, decision making /path planning, and automatic control. To accomplish the functionality of full autonomy, these three key modules need to cooperate, starting from the perception module. The AV's perception system uses a set of onboard sensors to detect and understand the surrounding driving environments containing both static and dynamic obstacles, such as roadside objects, road boundaries, moving pedestrians and vehicles, and traffic signs. Light Detection and Ranging (LiDAR) and cameras are two of the most essential and commonly used sensors for scene understanding in vehicle perception systems. Among them, the camera is a low-cost, small in size, and higher-resolution sensor providing 2D information like appearances about the environment in the format of RGB or gray-scale images. Cameras generally work well in clear weather conditions with good illumination. LiDAR is a remote sensing technology that utilizes pulsed laser beams to obtain 3D information of surrounding objects, including accurate distance measurements. Current LiDAR usually has lower resolution compared to cameras but can work under more variety of weather conditions. According to their pros and cons, the idea of fusing both sensors could leverage the advantages with overcoming their disadvantages. Moreover, having multiple multi-modal sensors allows for redundancy when facing failures from individual sensors. Therefore, the LiDAR-camera combination is an effective way for both object detection and range calculation for safe autonomous diving\cite{11}\cite{12}. 

The key part of a LiDAR-camera-based perception system is sensor fusion, which synthesizes inputs from the two types of sensors via coordinate transformation, data point mapping, and 3D scene reconstruction of the surroundings. Many researchers have contributed themselves to the development of LiDAR-Camera fusion algorithms. An interactive LiDAR to camera calibration toolbox to estimate the intrinsic and extrinsic transforms has been introduced in \cite{13}. The authors in \cite{14} address the common, yet challenging, LiDAR-camera semantic fusion problems, which are seldom approached in a wholly probabilistic manner. A novel open-source tool for extrinsic calibration of radar, camera, and LiDAR has been presented in \cite{15} with facilitating joint extrinsic calibration of all three sensing modalities for multiple measurements. \cite{16} presents a pipeline for extrinsic calibration of a ZED stereo camera with a Velodyne Puck LiDAR. This pipeline uses a novel 3D marker whose edges can be robustly detected both in the image and 3D point cloud. In another recent study towards extrinsic calibration of LiDAR-Camera fusion, an algorithm has been introduced in \cite{17} to estimate the similarity transformation between laser points and pixels using a checkerboard. 

Besides the conventional efforts to improve the accuracy of transformation matrices, learning-based approaches have also been exploited recently. A study on pedestrian classification used a deep-learning-based method taken from data from a monocular camera and a 3D LiDAR sensor, and compared early and late multi-modal sensor fusion approaches \cite{18}. In \cite{19}, another deep learning approach has been developed to carry out road detection by fusing LiDAR point clouds and camera images. An unstructured and sparse point cloud was firstly projected onto the camera image plane and then upsampled to obtain a set of dense 2D images encoding spatial information. 

Although many efforts have been put into the LiDAR-camera fusion research, the LiDAR-Camera mapping error is still pervasive in practice. In \cite{20}, fusion errors were measured by the distances between the actual chessboard 3D locations and the mapped LiDAR points, and then analyzed quantitatively using simulation and real test sequences to refine the transformation matrices further. Another study also reported LiDAR point projection errors on the images and tried to use a target-based 3D-LiDAR-to-camera calibration method to achieve a 50\% reduction in error values and 70\% reduction in variances \cite{21}. True positive, false positive, and false negative are proposed to use as main metrics for fusion accuracy evaluation \cite{22}. Compared with the studies proposing fusion algorithms to improve accuracy in simulation or static lab environments, there are much fewer studies analyzing the mapping errors in practice and focusing on minimizing their effects. In a real road environment, issues like time synchronization errors, hardware/software limitations, vibrations, high moving speed of the sensors and target objects, unreliable object detection, and others can all contribute to LiDAR-camera mapping errors. There are very limited papers focusing on these errors in practice and how to deal with them. 

The paper is organized in the following way. The problem statement and research objectives are discussed in Section II. Section III talks about the system calibration process. The probabilistic fusion process is introduced in Section IV. Section V illustrates the performance evaluation for the proposed method. Finally, the conclusions are presented in Section VI.

\section{Problem Statement and Research Objective}

In this research, we focus on the observed errors of LiDAR-Camera fusion from a practical view and propose a low-cost probabilistic algorithm to reduce the effects of the mapping errors on the fusion results. The research is motivated by two assumptions:

\begin{itemize}
    \item Although static sensor calibration can achieve very high fusion accuracy, the fusion of fast-moving small targets at long distance will still generate large mapping errors in the dynamic driving environment. This situation is more important with the increased popularity of micromobility vehicles like e-scooters. These tools tend to be small, fast, agile, and share the road with cars. 
    \item More advanced techniques like online learning and automatic calibration can potentially reduce the errors, with the cost of higher computational resources. This will limit the applications of such algorithms in wider range of embedded computing systems. A low-cost solution to the errors may satisfy the needs in many situations without intensive work on calibration. 
\end{itemize}

\begin{figure}[!htbp]
  \begin{center}
  \includegraphics[height = 1.6in, width = 3.3in]{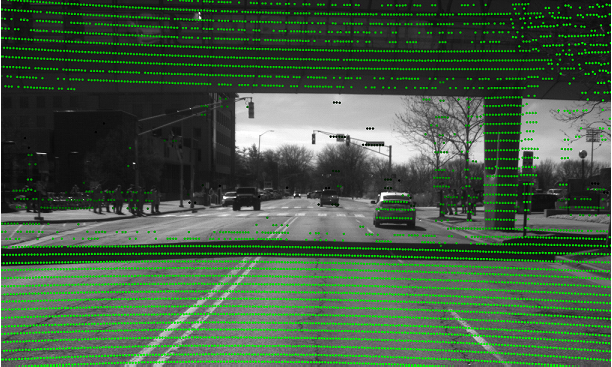}
  \caption{Example of an accurate LiDAR-Camera mapping result}\label{Long_diagram}
  \end{center}
\end{figure}

The goal of LiDAR-camera fusion is to map LiDAR cloud points onto the corresponding image pixels. In other words, this process tries to provide the values in the 3D LiDAR Cartesian coordinate system for a pixel in the 2D image coordinate system, which will then give any detected object in the image with a relative position from the vehicle. Accurate fusion results can be seen in Figure \ref{Long_diagram}, where the green dots on the image represent the mapped LiDAR dots on the pixels. The edges of the same objects from both sensors should align well, and the dots are well distributed across the object surface. 

\begin{figure}[!htbp]
  \begin{center}
  \includegraphics[height = 1.8in,width = 3.3in]{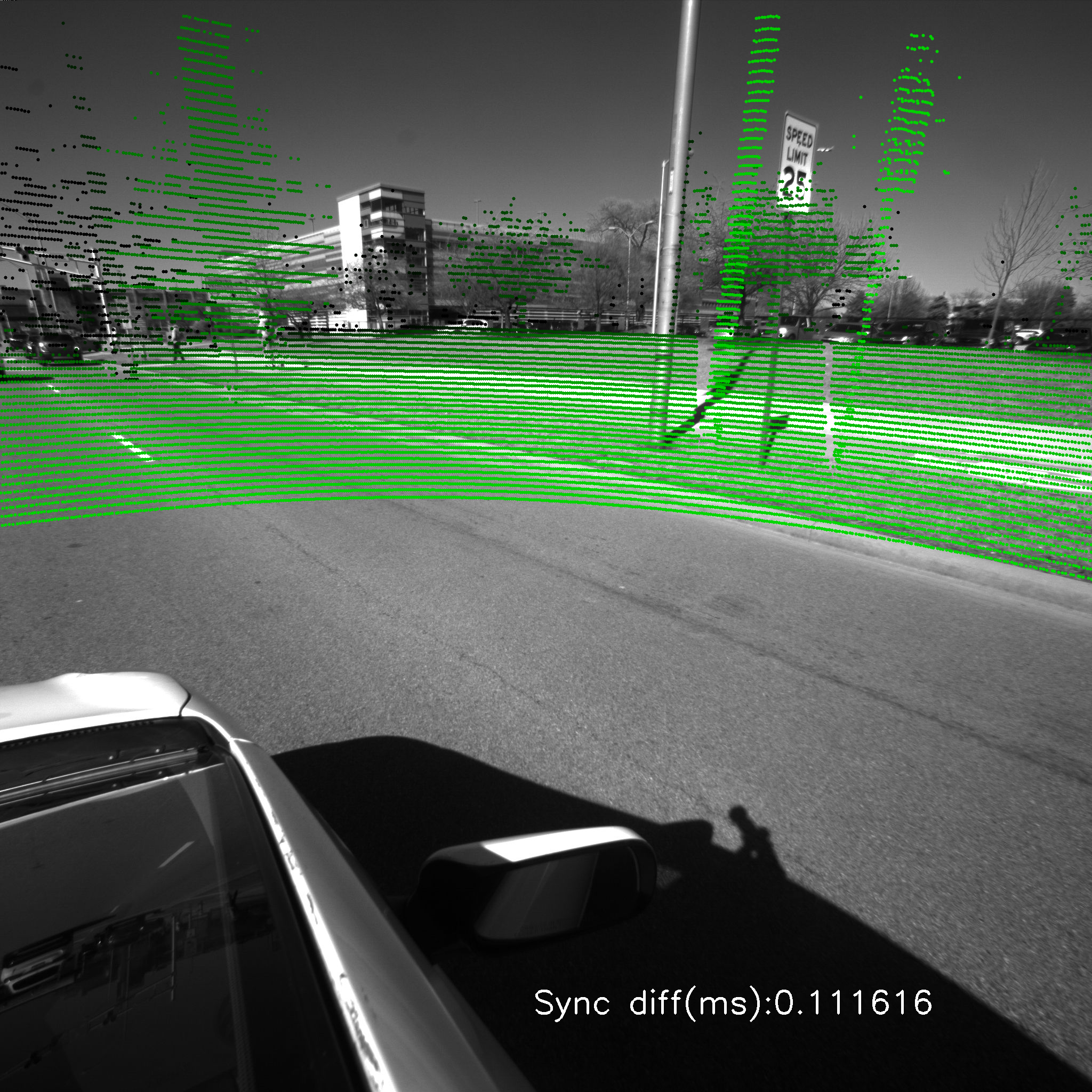}
  \caption{Example of an LiDAR-Camera mapping with errors}\label{Long_diagram_error}
  \end{center}
\end{figure}

In practice, two kinds of mapping errors can commonly happen. The first is that the background points got mapped into the area of interest (AOIs) as noises. The second one is the shifting of the projected LiDAR points to the correct image pixels. Figure \ref{Long_diagram_error} shows an example of wrongly mapped results from LiDAR-Camera fusion. Section 3 introduces the typical LiDAR-camera calibration process, making very accurate LiDAR-to-camera mapping results within the calibrated zone in a stable environment. In practice, imperfect time synchronization (due to sampling rate and saving time), fast motion, large angular accelerations, and locations beyond the calibrated zone may contribute to the errors.  

To address these inherent errors, the objective of this paper is to propose a multi-step probabilistic approach
to handle the errors when they are not avoidable. Data from larger AOIs are mapped first to deal with the shifting issues, and then a probabilistic filtering process is applied to remove noises and select only the correct mapping results. The fusion process is implemented in both car-based and wearable LiDAR-camera sensing systems, respectively, towards the interactions between car and e-scooter riders. We choose e-scooter rider for the experiments, considering this new type of micro-mobility tool as one of the most challenging road objects for AV perception systems due to its small size and fast movement. Comparisons are made among fusion outputs, baseline data, and ground-truth measurements to evaluate the performance of the proposed fusion method. 
The main contributions of the paper are summarized as below:\\
\begin{itemize}
    \item A approach is proposed to improve the fusion results when LiDAR-camera mapping errors are not avoidable in practice. 
    \item The proposed method requires less computational resources and can supplement the efforts to improve mapping accuracy. 
    \item The KL-divergence-based shape descriptor can capture geometric information from LiDAR point distributions efficiently with simplified similarity calculation.
    \item The proposed calibration and fusion method can support the detection and localization of micromobility vehicles like e-scooters efficiently. 
\end{itemize}

\section{Probabilistic Fusion Process}

\begin{figure}[!htbp]
  \begin{center}
  \includegraphics[height = 2.5in,width = 3.3in]{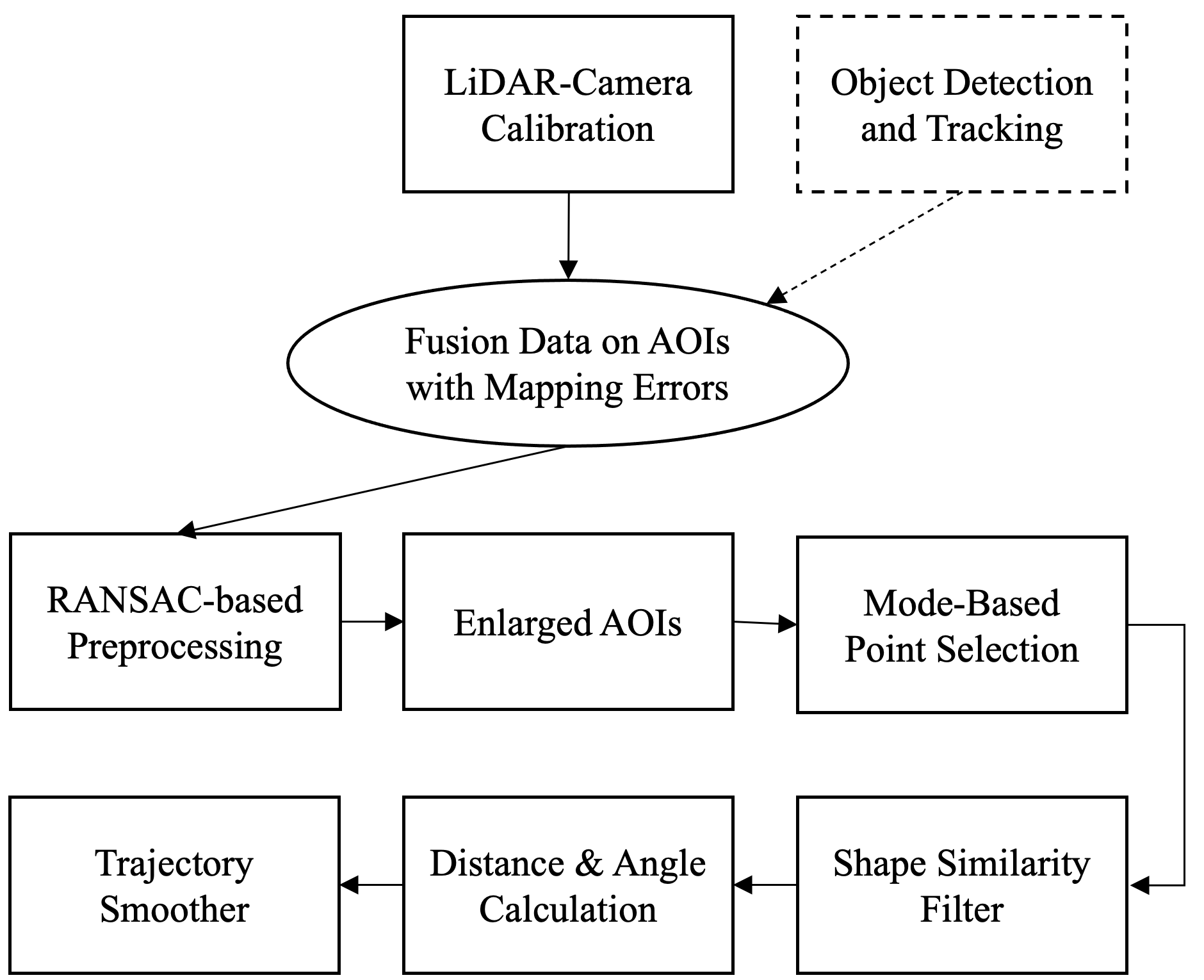}
  \caption{The proposed multi-step probabilistic fusion process that is robust to mapping errors}\label{Probabilistic Fusion Process}
  \end{center}
\end{figure}

The proposed probabilistic fusion process targets integrating the data from a monochrome camera and a 3D LiDAR to localize the surrounding objects of interest and track their relative trajectories related to the ego vehicle. The multistep process is illustrated in figure \ref{Probabilistic Fusion Process}. 

After the calibration of the LiDAR and camera pair, the collected LiDAR point cloud will be mapped onto the image pixels. The AOIs will be identified frame by frame via computer-vision-based object detection and tracking algorithms. Because these computer vision algorithms are not the main focus of this study, they will not be discussed. Combining the detection and tracking results with the mapped data will generate fusion results on AOIs. If everything is correct, the LiDAR points in the AOIs can be used to localize the corresponding objects via their 3D coordinates. As discussed earlier, two errors might occur in the outputs at this step, preventing direct localization of the objects. Thus, we propose a six-step pipeline to localize the objects with potential errors efficiently. The following subsections will talk about these steps in detail. 

Moreover, we formulate the problem as the following. Without loss of generality, we ignore the continuity of an object. In other words, every pixel belongs to exactly one object.   We can define the image as $\mathcal{X}$: $$ \mathcal{X} =: \{(l,w)
\},$$ 
$$\forall \, (l,w) \, \in \mathcal{X}, \,\exists \, o_k \, and \, c_k \,\subset\, \mathcal{X} \, s.t. (l,w)\, \in o_k \, and \, o_k \subset c_k,  $$ where $o_k$ is the set of pixels belong to the object $k$ and $c_k$ is the set of pixels identified by the bounding box from a CV algorithm corresponding to the object $k$. 

Similarly, we can define $\mathcal{Y}$ as the LiDAR point cloud: $$ \mathcal{Y} =: \{(x,y,z)\}, $$ $$\forall \, (x,y,z) \, \in \mathcal{Y}, \,\exists \, p_k \subset\, \mathcal{Y} \, s.t. (x,y,z)\, \in p_k \, $$ where $p_k$ is the set of LiDAR points belong to the object $k$. 


A mapping $\mathbf{M}:\mathbf{Y}\rightarrow\mathbf{X}$ maps a LiDAR triplet $(x,y,z)$ to a pixel $(l,w)$. In ideal conditions, 

\begin{equation}
    \mathbf{M}(x,y,z) = (l,w)
\end{equation}

\begin{equation}
    \mathbf{M^{-1}}(l,w) = (x,y,z)
\end{equation}

Assuming $\epsilon_l$ and $\epsilon_w$ are the mapping errors, we have
\begin{equation}
    \mathbf{M}(x,y,z) = (l,w) + (\epsilon_l,\epsilon_w)
\end{equation}

Given the object of interest $k^*$, we propose the algorithm $G(.)$ to select more LiDAR points reflected by the object $k^*$ that are mapped in the bounding box $c_{k^*}$, s.t.

\begin{equation}
    m_{k^*}= G(\mathbf{M^{-1}}(c_{k^*}))
\end{equation}

\begin{equation}
    Pr([|p_{k^*}|-\sum_{(x,y,z)\in m_{k^*}}f(x,y,z,k^*)] < t_1) >= t_2
\end{equation}
where $$f(x,y,z,k) = \mathbf{1}_{\{(x,y,z)\in p_k\}},$$ 
and $t_1$ and $t_2$ are predefined thresholds.

\subsection{System Calibration}\label{SectionCalibration}

\begin{figure}[!htbp]
  \begin{center}
  \includegraphics[height = 1in,width = 3.3in]{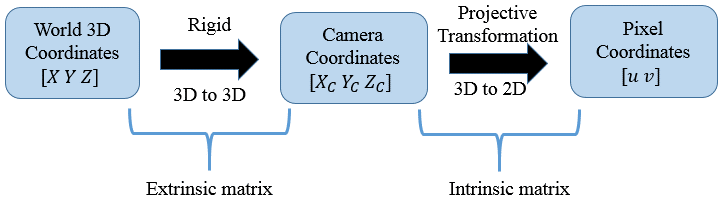}
  \caption{Process of system calibration.}\label{Calibration_diagram}
  \end{center}
\end{figure}

In order to map LiDAR data to the images, the LiDAR-camera system needs first to be calibrated. In this study, we follow the standard calibration process for LiDAR-camera fusion, primarily calculating three matrices, namely the “intrinsic camera matrix,” the “distortion parameters,” and the “extrinsic transformation matrix”\cite{23}\cite{24}. The camera matrix contains data representing camera parameters, such as focus, shear, and image width and height. The distortion coefficients represent the corrections due to lens distortion, such as for the fish-eye lens, which are assumed to be zero for lenses with normal field-of-view (FOV). In this study, we will assume the distortion of the images is minimal and will only consider two components of intrinsic camera matrix and extrinsic matrix for system calibration. The process is depicted in Figure \ref{Calibration_diagram}. In the figure, from left to right, the 3D LiDAR point cloud data can be converted to 3D camera coordinates using the extrinsic matrix. 
The intrinsic matrix will then be applied to map the 3D camera coordinates to the 2D pixel coordinates in the image plane. 



A three-step process is applied to calculate these matrices. The intrinsic camera matrix is calculated using the MATLAB Calibration Toolbox. The process of obtaining the extrinsic matrix started from calculating an initial extrinsic matrix using the Levenberg-Marquardt algorithm, which is also known as the damped least-squares method for solving $AX=B$ matrix equations. The algorithm is implemented using the OpenCV SolvePnP function. After getting this initial matrix, a Python script was used to fine-tune the extrinsic matrix manually.

The sensor fusion process can be shown in the following intrinsic-extrinsic matrix equation \ref{FusionEquation}, which governs LiDAR to camera coordinate transformation in the homogeneous representation. 

\begin{equation}\label{FusionEquation}
 \begin{bmatrix} u \\  v \\ 1 \end{bmatrix}
 =
  \begin{bmatrix}
   \text{f}_x & 0 & \text{o}_x & 0\\
   0 & \text{f}_y & \text{o}_y & 0 \\
   0 & 0 & 0 & 1
   \end{bmatrix}
   \begin{bmatrix} 
   \text{r}_{11} & \text{r}_{12} & \text{r}_{13} & \text{t}_x\\
    \text{r}_{21} & \text{r}_{22} & \text{r}_{23} & \text{t}_y \\ 
   \text{r}_{31} & \text{r}_{32} & \text{r}_{33} & \text{t}_z \\
    0 & 0 & 0 & 1
   \end{bmatrix}
   \begin{bmatrix} X \\  Y \\ Z \\ 1 \end{bmatrix}
\end{equation}

\noindent where u and v are the pixel coordinates in the image plane, and the X, Y, and Z are the coordinates from LiDAR measurements. The first matrix is a 3×4 camera intrinsic matrix. $f_x$ and $f_y$ are the focal lengths, which are expressed in pixel units, and the principal point ($o_x$,$o_y$), usually means the image center. The second one is a 4×4 extrinsic matrix, which is composed of a 3×3 rotation matrix R and a 3×1 translation vector T in the equation.

In ideal conditions, LiDAR point clouds can be accurately mapped onto the image by directly applying equation \ref{FusionEquation} after the calibration process. This is the mapping process of $\mathbf{M}$ defined earlier. However, different factors will result into mapping errors. From the next step, we use six steps to reduce the effects of such errors on object detection and localization.

\subsection{RANSAC-based Prepossessing}
Towards the mapping results with potential errors, the first step is to clean well-organized noise. Random Sample Consensus (RANSAC) is originally used to randomly sample data to construct a best-fit estimate, which can be used here for fitting the ground plane. To run RANSAC more efficiently, we also set up a range limit, and LiDAR points that exceed the limit are removed directly. The detailed process of applying the RANSAC algorithm in this paper is shown as below:\\
\begin{itemize}
 \item Step 1 - Set the map boundary as 70 m long and 30 m wide for the ego object and extract all point cloud data in the boundary
 \item Step 2 - The normal direction of the ground plane should generally point upward along the Z-axis as (0,0,1)
 \item Step 3 - Set the maximum point to plane distance ($\delta$ = 0.2m) for plane fitting
 \item Step 4 - The number of trials N is defined as
 \begin{equation}
     N = \frac{ln(1 - p)}{ln[1 - (1 - \epsilon)^n]}
 \end{equation}
 where p = 0.99 is the probability that at least one of the N trials will be free of outliers in the estimated ground plane. $\epsilon$ = 0.2 is the probability that a randomly chosen point is an outlier, and n = 6 is the number of 3D points used in each trial to compute the ground plane.
\item Step 5 - The minimum value for the size of the inlier set for it to be acceptable is determined as \\
\begin{equation}
 P = (1 - \epsilon)\Sigma
\end{equation}
where $\Sigma$ is the total number of 3D LiDAR points

\item Step 6 - Only the point clouds that are not in the ground plane are selected. 
\end{itemize}

\begin{figure}[!htbp]
  \begin{center}
  \includegraphics[height = 1.3in,width = 3.3in]{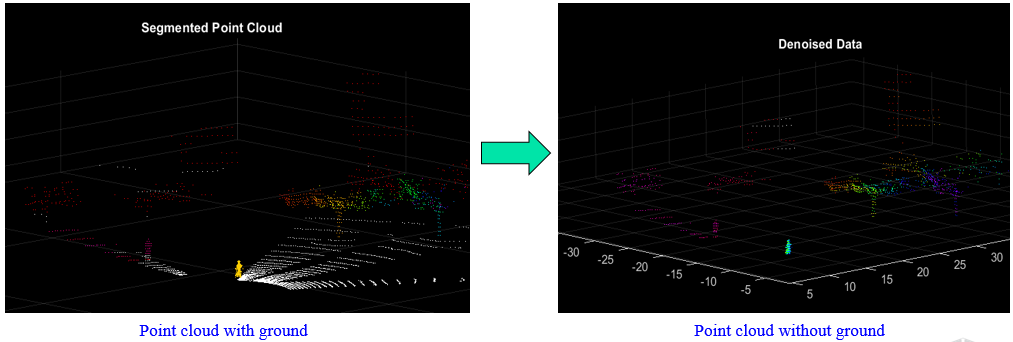}
  \caption{Results of removing ground points using RANSAC.}\label{RANSAC}
  \end{center}
\end{figure}

After the prepossessing step, a considerable portion of noises are eliminated.

\begin{equation}
    \mathcal{Y'}=RANSAC(\mathcal{Y})=\mathcal{Y}\setminus m_g
\end{equation}

\noindent where $m_g$ is the set of LiDAR points of all the ground, and $\mathcal{X'}$ is the set of all LiDAR points excluding ground points. 
As shown in Figure \ref{RANSAC}, the ground point cloud data (white points shown in the left image) can be deleted in the collected LiDAR data. In the filtered 3D plot, objects such as e-scooter riders, pedestrians, and trees can be observed very clearly (shown in the right image).     

\subsection{Enlarged AOIs}
After mapping LiDAR points to the detected objects and removing ground reflections using RANSAC-based preprocessing, two major issues explain the remaining data noise: 
\begin{itemize}
    \item Due to various system and algorithm limitations, the LiDAR points reflected from the target object may be mapped away from the center of the AOIs or even out of the detected AOIs. If only the mapped data within the AOIs are considered as inputs, the localization algorithm may use only partial or none correct LiDAR points. 
    \item The mapped LiDAR data on AOIs may contain reflections from other objects because of mapping offsets and inaccurate outputs from the object detection algorithms. This will bring the background points (mainly from other objects in the view) into the localization calculation of the target object, if not well addressed.
\end{itemize}

 \begin{figure}[!htbp]
  \begin{center}
  \includegraphics[height = 2in,width = 2.2in]{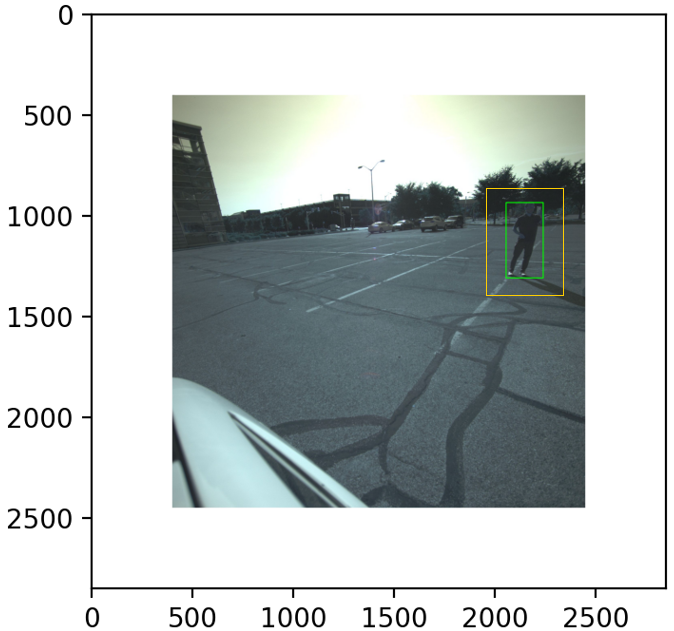}
  \caption{Sample of a enlarged AOI}\label{enlarge}
  \end{center}
\end{figure}

We address the first issue by enlarging the AOIs (Fig.\ref{enlarge}). The logic is straightforward that larger AOIs will cover more adjacent points to ensure the inclusion of more LiDAR points in the target object. Considering the random directions of possible offsets, the enlargement also goes to all 4 directions, in proportion to the size of the bounding box. The exact ratio is a hyperparameter to be selected based on the system and application. In our case, 100\% enlargement was implemented for left and right directions.
\begin{equation}
    c_{k^*}^{'}=Large(c_{k^*})
\end{equation}

\noindent where $c_{k^*}^{'}$ is the set of all pixels in the enlarged AOI $k^*$, and $Large(c_{k^*})$ is the AOI enlargement process. 

Since AOI enlargement exacerbates the issue of adding more noise from the background or irrelevant objects, there is a trade-off between the influence from the first and the second issue. Next steps will focus on the issues of remaining noise. 

\subsection{Mode-Based Clustering}\label{Model-Clustering}

\begin{figure}[!htbp]
  \begin{center}
  \includegraphics[height = 2in,width = 3.3in]{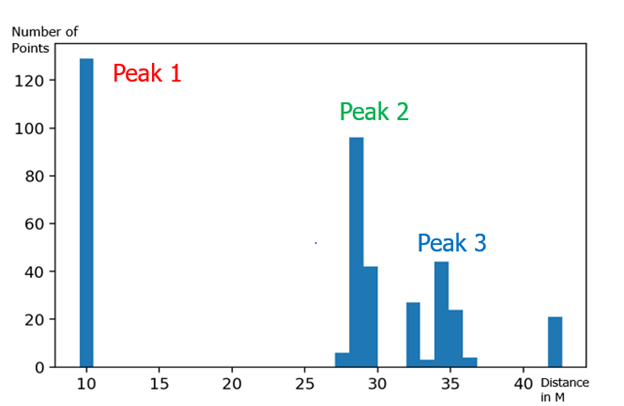}
  \caption{Distance distribution}\label{dist}
  \end{center}
\end{figure}

In the previous steps, enlarged AOIs will ensure the inclusion of correct LiDAR points from the target object, with the risks to include more noises from other objects. These noises tend to be isolated clusters because the continued reflections from the ground have been removed in the preprocessing stage. We devise an method based on three assumptions to fetch the correct LiDAR points in this step. 
\begin{enumerate}
    \item The first assumption is that the expanded AOI includes most of the correct LiDAR points from the target object, which assures that enough LiDAR points in the AOI are from the target object. 
    \item Secondly, if the AOI includes multiple objects, the target object is one of the largest. This assumption ensures successful generation of data characteristics. 
    \item Lastly, the LiDAR points from one object should have similar distance readings from the LiDAR, assuming the object is a continuum. 
\end{enumerate}

Then, we can create the distance distribution of all LiDAR points in the augmented AOI under a certain granularity level. The granularity level is a hyperparameter based on the class of the target object (for example, vehicle should have a larger granularity than pedestrian). To further decide the bin positions, we cluster the points using K-Means to find high-density distance centers (which are assumed to be the distance center of objects), and use them as the center of bins. Since in later steps, bin-based clusters are used to represent each object, a miss-aligned bin position can significantly affect the correct point selection. When key bins are positioned, the rest of the bins will be fulfilled sequentially following the hyper-tuned granularity level. 

After determining the bin positions and sizes, the distribution of all measured distances from LiDAR points in the enlarged AOIs can be created. A sample distance distribution with three distance centers (peaks) is shown in Figure \ref{dist}. 
Such a distance distribution clusters the distance-similar points together. Furthermore, based on the second assumption above, we consider the histogram columns with higher densities as potential candidate clusters for the target object. If there is only one concentrated bar in the distance distribution, we use LiDAR points in this cluster for the target object. There are two hyper-parameters to further filter the candidate clusters for situations with more than one peak: 
\begin{itemize}
    \item The minimum number of points required for a peak to be selected.
    \item The proportion between the number of points in the selected peak and the highest peak.
\end{itemize}

These two hyper-parameters should be tuned based on the performance in the individual applications. In Figure. \ref{dist}, by applying different values for the filter hyper-parameters, two or three from the highest identified peaks may be qualified. 

\begin{equation}
Mode(\mathbf{M}^{-1}(c_{k^*}^{'}))=(m_1,m_2,\cdots,m_n)
\end{equation}

\noindent where $m_1$,$m_2$,$\cdots$,$m_n$ are qualified candidate clusters of LiDAR points, and $Mode(.)$ is the mode-based clustering.

\subsection{Shape Similarity Filter}

If only one cluster was selected from the mode-based clustering process, then the data in the cluster can be used for further steps; otherwise, one shall be selected from multiple clusters for the target object. This is achieved by calculating shape similarity among each cluster with the target object. 

The logic is that the target object should have a similar shape as the type it belongs to. The type of the object is known from the object detection algorithm. The rest of the LiDAR points in the enlarged AOI are either from noise, other types of objects, or a partial shape of the same type of object (by assuming that the object detection algorithm will output a accurate bounding box surrounding the target object). Thus, the cluster of LiDAR points that has the highest similarity with the targeted object type should be selected.

\subsubsection{Shape Descriptor}

We use shape descriptor to capture geometric information and simplify the shape representation, the following three steps:
\begin{enumerate}
    \item The 2D mapping of LiDAR points from each candidate cluster is divided into nine sections evenly.
    \item The percentage of points in each section is calculated.
    \item The nine percentages form a vector to represent the shape information of the candidate LiDAR point cluster. 
\end{enumerate}

\begin{figure}[!htbp]
  \begin{center}
  \includegraphics[height = 3in,width = 3.3in]{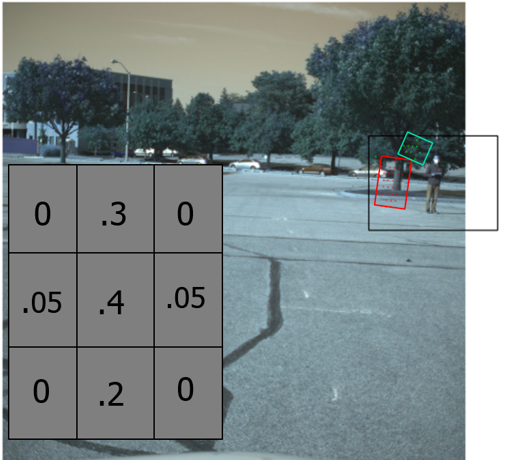}
  \caption{Shape similarity and rotation demo}\label{rotation}
  \end{center}
\end{figure}

The bottom-left table in Figure \ref{rotation} shows an example of the generated nine percentages, which can be flattened to form the shape descriptor vector.. 

\subsubsection{Benchmark Shapes for Different Objects}

To make a comparison with the target object type, benchmark shape descriptors for each type of targeted objects need to be created. These distributions (vectors of the shape descriptor) can be learned by processing and averaging a list of manually selected objects for each type of interest (vehicle, pedestrians, e-scooter riders, etc.). The number of objects to be processed to create the benchmark can be tuned as another hyperparameter. 

\subsubsection{Shape Rotation}

The proposed shape descriptor is invariant to the number of LiDAR points or the size of the bounding box, but is sensitive to rotation. Even for the same object, the rotated LiDAR points will represent quite different shape descriptor vectors, which may happen when the vehicle bumps or vibrates to cause large angular movements of LiDAR and camera in real road environment. 

To deal with the rotation issue, we use a constrained PCA (Principal Component Analysis) to find the angle between the object's main direction and the ground and rotate the selected LiDAR points accordingly. The application constrains the maximum rotation angle to be less than 40 degrees. Since the PCA could identify the basis with maximum variation, the shapes of objects are rotated along with it. 
The constrained PCA works based on the assumptions that the target object is standing straight on the ground, where the first component's direction is along with the object's body. 




\subsubsection{Similarity Calculation}

Given the rotated shape descriptor vectors, the distances or differences among them can be calculated using Kullback–Leibler (KL) divergence and project it onto a probability measure through a modified sigmoid function shown in Equation \ref{KLequation} below. In the equation, P and Q are two discrete probability distributions with the same probability space $x$; and in this case, they are two rotated shape descriptor vectors. 

\begin{equation}
D_{\mathrm{KL}}(P \| Q)=\sum_{x \in \mathcal{X}} P(x) \log \left(\frac{P(x)}{Q(x)}\right)\label{KLequation}
\end{equation}

After calculating the KL-divergence scores among all candidate LiDAR point clusters with the benchmark shape for the target object's type, the cluster with the highest score is identified as the object, as in equation \ref{ClusterSelection}. 

\begin{equation}
\begin{split}
m^*= \argmax_{m_1,m_2,\cdots,m_n}(D_{\mathrm{KL}}(Shape(m_1)\| S^*),\\D_{\mathrm{KL}}(Shape(m_2)\| S^*),\cdots,D_{\mathrm{KL}}(Shape(m_n)\| S^*))\label{ClusterSelection}
\end{split}
\end{equation}

\noindent where $shape(.)$ is the process of calculating shape distribution, $m^*$ is the set of LiDAR points we select corresponding to object ${k^*}$, and $S^*$ is the benchmark shape distribution.

For the three peaks color-coded in Figure \ref{dist}, the following Figure \ref{shape} shows that they are from the target pedestrian shown in red (LiDAR data mapped to an offset), the two background vehicles are shown in green and blue, respectively. The shape similarity scores for the three clusters towards the object type of pedestrian are calculated and presented as the numbers in the upper-left corner in Figure \ref{shape}. For each row, the first number is the similarity score (confidence level) between the cluster and pedestrian shape before rotation, the middle number is the distance of the LiDAR point cluster, and the last number is the similarity score after rotation. Based on the outputs, the cluster shown in red will be identified as the target, with all points belong to the pedestrian correctly selected. 

\begin{figure}[!htbp]
  \begin{center}
  \includegraphics[height = 2in,width = 3.3in]{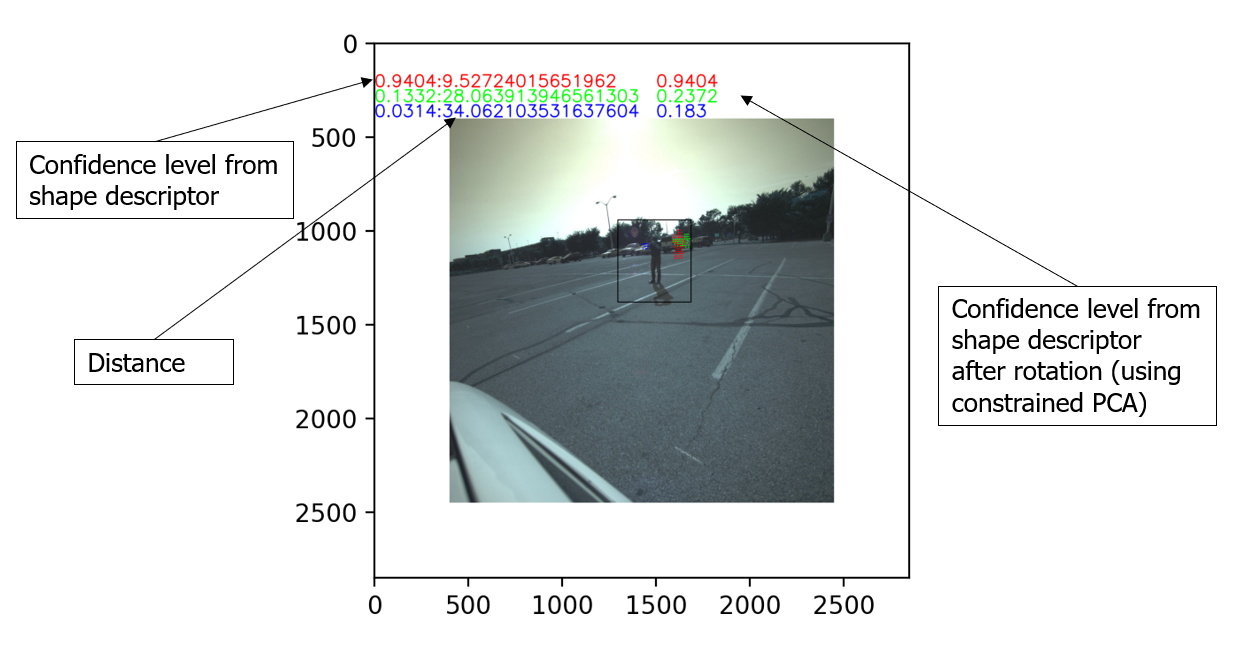}
  \caption{Distance calculation demo. (The first number in each row is the confidence level of the human shape before rotated. The second number is the distance. The third number is confidence level after rotation.)  }\label{shape}
  \end{center}
\end{figure}

\subsection{Distance \& Angle Calculation}
The main goal of the fusion process is to localize the target object by calculating its angle and distance. After the previous processes, we identified LiDAR points from the selected cluster. Since we have a specific granularity in the distance distribution before selecting the bin-based cluster, the distance and angle error are bounded within the granularity level. There are two strategies to calculate the distance and angle from these data:
\begin{enumerate}
    \item Averaging across all the points in the clustered LiDAR cloud
    \item Selecting one representative LiDAR point for the object
\end{enumerate}

We believe that both methods may generate similar results in most cases. However, in order to increase the robustness against possible distribution, skewness and outlier issues when the bin-width is relatively large, we select the point with the median distance among all the LiDAR points in the cluster to represent the object. This method can ensure an accurate estimation as long as more than 50\% of points in the cluster belong to the target object.  

\begin{figure}[!htbp]
  \begin{center}
  \includegraphics[height = 2in,width = 3.3in]{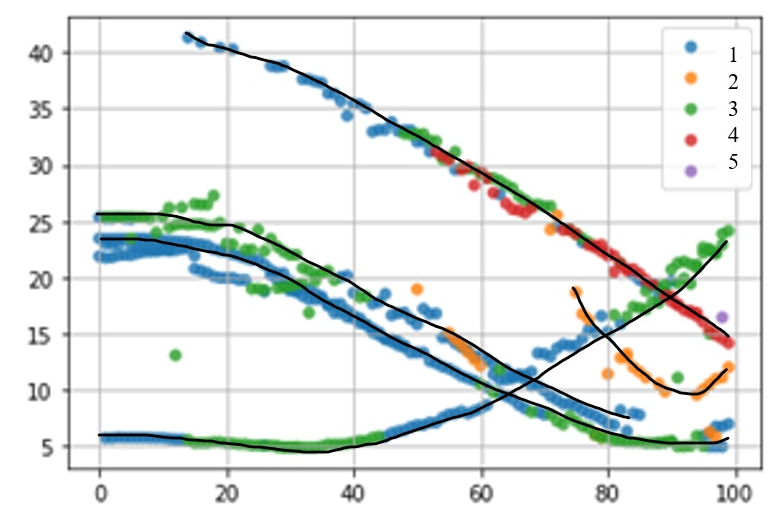}
  \caption{Multiple objects distance calculation. (Y-axis is distance, and X-axis is frame ID. Different colors are used to label the camera the objects are captured from. The black line is manually annotated distance for each object.) }\label{distance}
  \end{center}
\end{figure}

Both distance and angle calculations are based on this representation point. The distance of an object is simply the Euclidean distance calculated based on the representation point's X and Y values. Since the LiDAR is usually equipped on the top of a car, including z-dimension in the calculation might overestimate the distance. The angle between X-axis in the LiDAR's coordinate system and the representation point's vector in the X-Y plane is used to identify the object's orientation. 

Figure \ref{distance} shows an example of distance calculation for five objects continuously for 100 frames using multiple cameras. These cameras face different directions with overlap and are fused with the same LiDAR using the proposed probabilistic fusion process. Cameras and the LiDAR are synchronized. When multiple objects move in the scene, the LiDAR can always capture them, and different cameras may capture different objects at different times, together or individually. Then frame-by-frame, the distances of different objects calculated from different cameras are drawn in the plot. Different colors are corresponding to the cameras which capture the objects for that calculation. The curves align continuously and smoothly, and the same objects' distances captured by different cameras are consistent. We use Figure \ref{distance} to give a sense of our method's performance and will discuss the details in a later section.

\subsection{Trajectory Smoother}

The proposed fusion process is based on the assumptions mentioned above in section \ref{Model-Clustering}. The algorithm can localize the target object accurately when a small number of errors exist in the LiDAR and camera mapping results. However, when a large proportion of the LiDAR points in the enlarged AOIs are not from the target object, the distance and angle calculations are problematic. These errors need to be addressed in this step.

Supported by the object detection and tracking, consecutive localization can be completed frame by frame to form the trajectory of the target object, as illustrated in Figure \ref{distance}. A trajectory smoother is implemented to detect the outliers as the localization errors, under two assumptions: 
\begin{itemize}
    \item Assumption 1: The target object's movement is continuous in three dimensions, given the sampling rate of at least 10 frames per second. 
    \item Assumption 2: The mapping errors are from the distance of other objects, which will result in a distance difference larger than the size of the object.
\end{itemize}

Then for any given duration, t\textsubscript{0} to t\textsubscript{1}, we use two phases of regressions with different orders to smooth the trajectory. The first phase uses three order-two polynomial regression models with RANSAC to detect outliers (wrong distance/angle calculations) in the trajectory, corresponding to the three dimensions in the Cartesian 3D space. After we excluded all the outliers, we conduct three order-three polynomial regressions to smooth the inliers and interpolate the missing values. Using two phases of regression is because higher degree-of-freedom regression models are easily affected by outliers. Therefore, we lower the polynomial regression's order in the outlier detection phase. A higher-order polynomial regression can capture the curvature of the trajectory more accurately given changing accelerations and thus is used after the outliers are removed.  

During the phase of outlier detection, RANSAC is applied iteratively to sub-sample random points and fit a regression model. After certain iterations, the regression model of the best fit will be considered as the correct model. The points whose deviation from the best fit model exceeds the threshold are classified as outliers. In this process, we set the threshold as twice the value of the standard deviation.

The independent variables of each regression model are X, Y, Z values from each dimension, respectively. 
A point will be identified as an outlier if one of the X, Y, and Z values exceeds the corresponding model's threshold. For example, if a noise LiDAR point is reflected from an object behind the target object, and both two objects are in front of the LiDAR, the smoother can not detect this noise point through X and Z values. But our schema will also check the Y value and find out that it is an outlier.  

The final outputs of this probabilistic fusion process will be a smoothed trajectory of target objects with minimum effect from LiDAR-camera mapping errors.

\section{Performance Evaluation Method}

The performance of the proposed probabilistic fusion process is evaluated with experiments. To thoroughly test the process, we developed two hardware systems with different sensors and ran them on different platforms to perform LiDAR-camera fusion. Then the experiments were conducted to collect sensing data and ground-truth data of interactions between a car and an e-scooter rider in a test track. The performance of the proposed fusion process is evaluated by comparing the processed outputs with ground-truth results. The detailed design of the system can be referenced by \cite{25}. 

\subsection{Data Collection Apparatus}

We developed two hardware systems to support the evaluation of the proposed LiDAR-camera fusion process. One is the car-based LiDAR-camera perception system, and the other one is the wearable LiDAR-camera perception system. The descriptions of these two systems will be introduced in the following subsections.

\subsubsection{System I: Car-based LiDAR-camera Sensing System}
The car-based sensing system includes six FLIR Grasshopper 3 cameras, one 64-beam 360-degree Ouster OS1 LiDAR, a Reach Emlid RTK (Real-time Kinematics) GPS, one IMU (Inertial Measurement Unit),  and a desktop computer. The desktop computer, GPS module (excluding antenna), LiDAR, amplifier, and inverter are all placed in the car's trunk.  The main goal of the car-based system is to cover 360 degrees around the vehicle for scene reconstruction and an accurate self-localization via GPS and IMU. 

\begin{figure}[!htbp]
  \begin{center}
  \includegraphics[height = 1.2in,width = 3.3in]{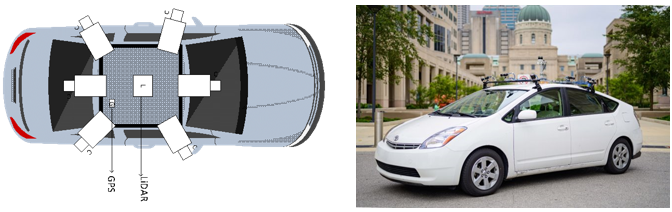}
  \caption{Overview of the car-based perception system with sensors.}\label{Car-based-System}
  \end{center}
\end{figure}

Four cameras are placed at the corners on the top of the vehicle, and two cameras are placed at the center, one front-facing, and one back-facing. The corner cameras are equipped with wide-angle (96 degrees) lenses, and the cameras at the center are equipped with 40-degree lenses. The LiDAR is placed at the center of the car's roof to provide maximum coverage around the vehicle. Figure \ref{Car-based-System} illustrates the overview of the car-based perception system with sensors. 

All sensors are connected to the PC and operated by the PC using ROS (Robot Operating System), which provides the flexibility to control and process multiple sensors simultaneously. Furthermore, hardware-based synchronizations at 10 FPS (frames per second) are achieved for LiDAR and cameras by sending a pulse sequence to the camera's sync port, and the GPS/IMU were continuously synchronized using the recorded UNIX timestamps from ROS bag files. Every rotation of the LiDAR (master) is synchronized with every camera capture frame (slave). 

\subsubsection{System II: Wearable LiDAR-camera Sensing System}

Unlike the car-based system, the wearable system cannot achieve a 360-degree view due to the occlusion of the human body. The cameras have to be carefully placed to cover the maximum available field of view in one direction (180-200 degrees). The wearable LiDAR-camera perception system consists of three USB cameras, a 3D LiDAR with an integrated IMU, an RTK GPS unit, and an NVIDIA Jetson Tx2 development computer board for data collection. 

\begin{figure}[!htbp]
  \begin{center}
  \includegraphics[height = 1.8in,width = 3in]{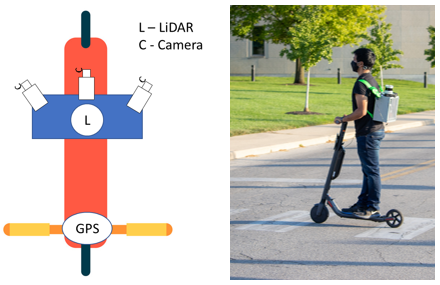}
  \caption{Overview of the Wearable LiDAR-camera perception system with sensors.}\label{Scooter-based-System}
  \end{center}
\end{figure}

In this study, we ask a e-scooter rider to wear the wearable system to capture it's own locations while also sense the car during interactions. As shown in Figure \ref{Scooter-based-System}, the system can be either in the front or the back of the rider. The LiDAR stays in the middle, and two cameras are placed on the corners and one in the center. The GPS antenna is placed in the front of the e-scooter while the system is back facing and on the rider’s helmet when front-facing. This design is because that the GPS needs to be at least 50cm away from other electronics. Three cameras cover 200 degrees of view in both front and back-facing configurations. The software architecture of the wearable system is similar to that of the car-based system. The main difference is that all the sensors are synchronized with the UNIX timestamps on the bag files.

\subsection{Experiment Design for Evaluation Data Collection}\label{SectionExperimentDesign}

\begin{figure}[!htbp]
  \begin{center}
  \includegraphics[height = 2.2in,width = 3in]{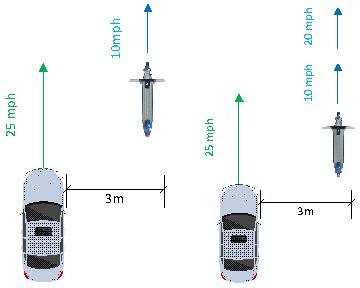}
  \caption{Designed scenario 1 (left) and scenario 2 (right).}\label{designedScenarios}
  \end{center}
\end{figure}

To better evaluate the fusion process, one experiment is designed using these two different hardware systems. The wearable system is equipped by a researcher who rides an e-scooter in an empty parking lot with the experiment vehicle installed with another set of the perception system. Both the experiment vehicle and the e-scooter are equipped with RTK (real-time kinematic) GPS sensors supported by multiple base stations, providing the ground-truth location to an accuracy of 10 centimeters at 5 frames per second (sampling time is 0.2 sec). The e-scooter and the car run in parallel during the experiment, as shown in Figure \ref{designedScenarios}. This scenario represents the situation that the vehicle overtakes the e-scooter from the left side in a parallel path on a straight road with the same driving direction. The lateral offset is around 3m, and the moving speed is about 25 mph for the vehicle and 10-20 mph for the e-scooter, respectively. The target object of the vehicle's perception system is the e-scooter rider, and the target object of the wearable perception system is the experiment vehicle. There are pedestrians, parked vehicles, trees, light posts, and buildings in the background at different distances. 

Two trials of each scenario were completed during the experiments. In the world coordinate system, both the car and the e-scooter rider move from south to north along the y-direction. The distance between them is along the x-direction. For the car-based perception system, the collected time duration after post-processing is about 5.2 secs with driving distances of approximately 65 m and 38m for the car and e-scooter, respectively. For the wearable system, the collected time duration after post-processing is about 5.8 secs with driving distances of approximately 60 m and 30m for car and e-scooter, respectively. 

\subsection{Measurements of Mapping Errors}\label{MeasurementMappingError}

\begin{figure}[!htbp]
  \begin{center}
  \includegraphics[height = 1.3in,width = 3in]{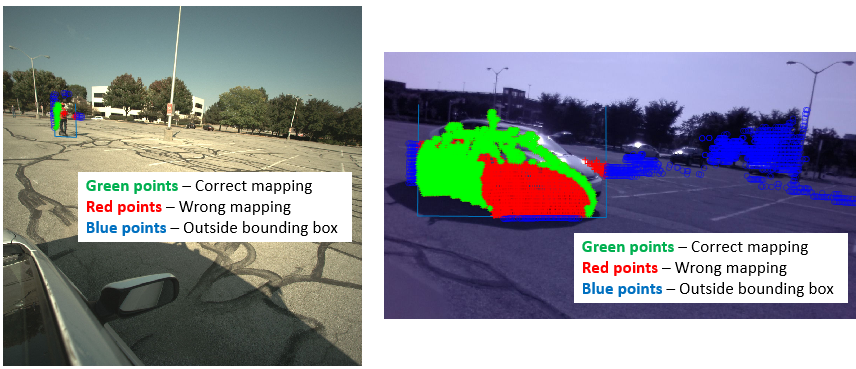}
  \caption{TPR of checking the fusion accuracy in scenario 1 for vehicle-based system (left) and e-Scooter-based system (right) using one camera.}\label{Scenario_2}
  \end{center}
\end{figure}

\begin{figure}[!htbp]
  \begin{center}
  \includegraphics[height = 1.3in,width = 3in]{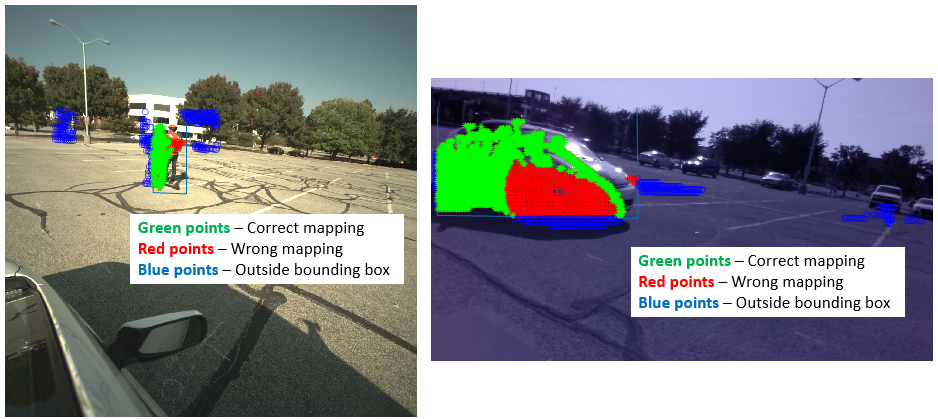}
  \caption{TPR of checking the fusion accuracy in scenario 2 for vehicle-based system (left) and e-Scooter-based system (right) using one camera.}\label{Scenario_4}
  \end{center}
\end{figure}

The True Positive Rate (TPR)  was calculated for both the car-based and the wearable LiDAR-camera perception systems to estimate the mapping errors. When judging the mapping correctness, the calculation assumes a valid distance variation up to $15\%$ of the size of the vehicle and e-scooter, respectively. The mapped LiDAR point for the object is correct if the corresponding distance is in that distance range. For example, if the ground truth distance is 10m for the car, and the car size is 4.5m long, the mapped LiDAR points with distances from 9.325 (10 - 0.15*4.5)m to 10.675 (10 + 0.15*4.5)m are considered correct mapping. Otherwise, they are wrong mappings. Similarly, if the ground truth is 10m and the e-scooter size is 1.5m long, the mapped LiDAR points with distances in 9.775 (10 - 0.15*1.5)m to 10.225 (10 + 0.15*1.5)m are considered correct mapping. Otherwise, they are wrong mappings. Then, TPR can be calculated using the following formula:

\begin{equation}
    True Positive Rate(TPR) = \frac{TP}{(TP + TN)}
\end{equation}

\noindent where TP is the number of corrected mappings, and TN is the number of wrong mappings. The higher value of TPR, the better the fusion accuracy. 

The examples of mapping errors in the two experiment scenarios were illustrated in Figure \ref{Scenario_2} and Figure \ref{Scenario_4} for both vehicle-based and e-scooter-based wearable systems, respectively. In the figure, the green points mean corrected mapping (TP), the red points are the wrong mapping (TN), and the blue points indicate that the mapped points are outside the bounding box.  

\subsection{Measurements of Localization Errors}\label{MeasurementLocalizationError}

Another measurement used for fusion performance evaluation is the localization error. The calculation of localization errors for both car-based and wearable perception systems is through comparing the estimated localization of the target via the proposed fusion process and the ground-truth localization using the installed RTK GPS module. The target localization can be defined as the coordinates (X, Y) in the perception system's coordinate system, with the origin on the corresponding car or e-scooter rider. Then, the localization errors in terms of localization X and localization Y can be calculated using Mean Absolute Error (MAE) as the following:

\begin{equation}\label{MAEx}
   MAE_{\mathrm{X}}= \frac{\sum_{i}^{N_x}\left|X_i - \hat{X_i}\right|}{N_x}
\end{equation}

\begin{equation}\label{MAEy}
   MAE_{\mathrm{Y}}= \frac{\sum_{i}^{N_y}\left|Y_i - \hat{Y_i}\right|}{N_y}
\end{equation}

\noindent where MAE means the mean absolute error and the subscript X indicates the coordinate X in the localization of the target. $X_i$ is the ground-truth coordinate X value of the target at the i-th timestamp in the scenario, and the $\hat{X_i}$ is the estimated coordinate X value after applying the proposed probabilistic fusion process at the i-th timestamp. $N_x$ is the total number of X points. For the MAE computation regarding the coordinate Y, the same definitions have been utilized.

\subsection{Data Analysis Method}

After conducting the experiments, collected data are processed following the proposed probabilistic fusion process. The two evaluation metrics described in sections \ref{MeasurementMappingError} and \ref{MeasurementLocalizationError} are calculated and compared at different levels to fully evaluate the performance.

\subsubsection{Mapping Accuracy}
The LiDAR-camera mapping accuracy is calculated at two levels:
\begin{itemize}
    \item Baseline mapping accuracy - The fusion result is firstly obtained when the static calibration matrices are directly applied following equation \ref{FusionEquation}. By following this typical fusion process, the mapping accuracy can be measured with the true positive rate in equation 9 based on the original ROI.
    
    \item Probabilistic fusion mapping accuracy - The mapping accuracy with larger ROI after utilizing the shape similarity filter can also be computed by following the same principle of true positive rate. 
\end{itemize}

\subsubsection{Localization Accuracy}
The object location is calculated at two levels:
\begin{itemize}
    \item Ground-truth location - since we have two GPS systems on each perception platform, the GPS coordinates of the ego object ($X_e$, $Y_e$) and target ($X_t$, $Y_t$) can be obtained. Thus, the ground-truth locations in terms of X and Y relative to the ego-object can be computed as ($X_t$ - $X_e$, $Y_t$ - $Y_e$).
    \item Probabilistic fusion localization accuracy - after applying the proposed probabilistic fusion process, object localization can be achieved through distance \& angle calculation. Some outliers can be further eliminated by applying the trajectory smoother. Eventually, a new set of more accurate target X and Y localization relative to the ego object will be obtained.
\end{itemize}

\subsubsection{Performance Evaluation}
The performance of the proposed probabilistic fusion process is evaluated using two hypotheses: 
\begin{itemize}
    \item Hypothesis 1: The proposed probabilistic fusion process can significantly improve the mapping accuracy from the baseline. 
    \item Hypothesis 2: The proposed probabilistic fusion process can achieve an average TPR that is larger than 0.5. When TPR is larger than 0.5, the process can guarantee that the fusion result is accurate. 
\end{itemize}

\section{Evaluation Results}
\subsection{Calibration Accuracy for Two Sensing Systems}

\begin{table}[pb]
\centering
\caption{Fusion accuracy under static calibration condition (TPR)}
\label{tab:calibration}
\begin{tabular}{l|l|l}
\hline
Pair Index & Car-based(\%) & Wearable(\%) \\ \hline
1 & 94.9 & 100 \\
2 & 100 & 100 \\
3 & 100 & 100 \\
4 & 100 & N/A\\
5 & 70 & N/A\\
6 & 100 & N/A\\
Average & 94.15 & 100 \\
\bottomrule
\end{tabular}
\end{table}

The two developed LiDAR-camera sensing systems are both calibrated following the process in section \ref{SectionCalibration}. The fusion accuracies under the static calibration condition for both systems are calculated using equation \ref{FusionEquation} and shown in the Table \ref{tab:calibration} below. 

There are six pairs of LiDAR-camera fusion for the car-based system. Out of all these pairs, four pairs have an calculated TPR of 100$\%$, with an average TPR of 94.15$\%$. Similarly, there are three pairs of LiDAR-camera fusions for the wearable system with all of them achieving an TPR of 100$\%$. This result proves that both systems are calibrated to achieve very high fusion accuracy in the static lab environment through the typical fusion process.


\subsection{Mapping Accuracy in Practice}

Although achieving high fusion accuracy in a static environment, the sensing systems encounter much higher errors in dynamic environments due to small synchronization offsets, high moving speeds, and vibrations. 

The average TPRs for both the car-based and wearable systems in the two experiment scenarios (as described in section \ref{SectionExperimentDesign}) is calculated across randomly sampled frames combining all camera-LiDAR pairs, with the results shown in the Baseline column in Table \ref{tab:performance_tpr}. The average TPRs for the car-based system is 31.72$\%$ in scenario 1 and 68.15$\%$ in scenario 2, and both are much lower than the TPRs in the lab environment. Similarly, the average TPRs for the wearable system are 48.76$\%$ and 55.77$\%$ in scenarios 1 and 2, respectively, which are also much lower than the 100$\%$ fusion accuracy in the static calibration. 

Overall, these baseline fusion accuracies contain pretty significant errors and are not good enough to localize the targets if directly used, especially for discontinuous or jumping frames with very low TPRs. A better sensing system may reduce these errors through advanced synchronization, more expensive hardware, and complicated and intensive calibration. However, it is still challenging to avoid mapping errors entirely. Considering that this study aims to solve this issue using a low-cost probabilistic fusion process, these errors provide a baseline. They are not further reduced through other methods. 

\begin{table}[h]
\centering
\caption{Paired two-sample t-test results for True Positive Rate (TPR). (Baseline and P-Fusion refer to the average TPR before and after the implementation of the algorithm, respectively.)}
\label{tab:performance_tpr}
\begin{tabular}{l|llllll}
\hline
System   & \multicolumn{1}{l|}{Sce} & \multicolumn{1}{l|}{Baseline(\%)} & \multicolumn{1}{l|}{P-Fusion (\%)} & \multicolumn{1}{l|}{\textit{t}} & \textit{p-val}  & N \\ \hline
Car-based      & 1                        & 31.72(25.30)                     & 65.02(19.01)                      & 8.01                            & \textless{}.001 & 10 \\
         & 2                        & 68.15(26.02)                     & 81.65(17.93)                      & 6.54                            & \textless{}.001 & 19 \\ \hline
Wearable & 1                        & 48.76(16.01)                     & 65.46(18.14)                      & 3.82                            & .006            & 6  \\
         & 2                        & 55.77(22.17)                     & 75.29(12.93)                      & 6.34                            & \textless{}.001 & 10 \\ \hline
\end{tabular}
\end{table}

\begin{figure}[!htbp]
  \begin{center}
  \includegraphics[height = 1.3in,width = 3in]{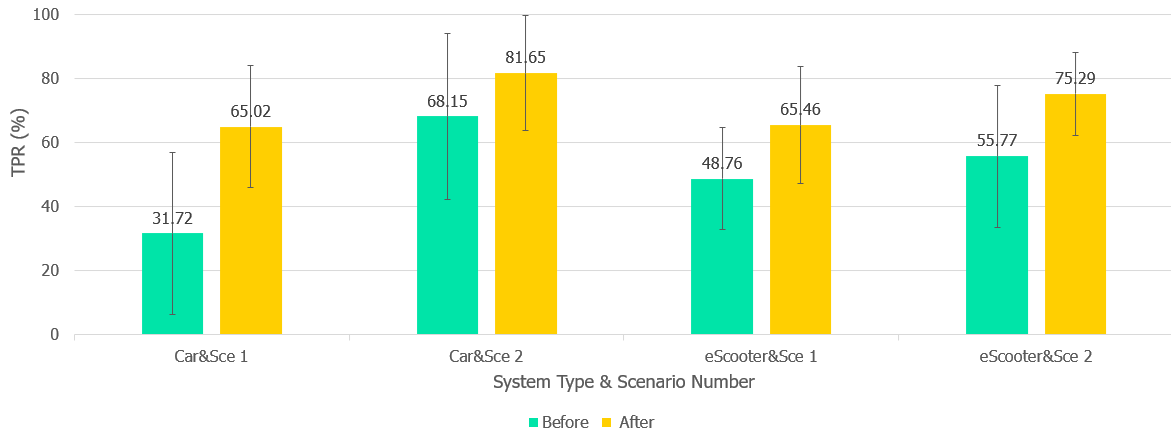}
  \caption{Average (std) TPR before or after the proposed algorithm for each system type and scenario.}\label{TPRbar}
  \end{center}
\end{figure}

For the same frames randomly sampled from the car-based and wearable systems during the two experiment scenarios, the proposed probabilistic fusion process was applied to improve the mapping accuracy. The average TPRs for these frames from the two systems are provided in the P-Fusion column in Table \ref{tab:performance_tpr}. As illustrated in Figure \ref{TPRbar}, the average TPRs are much improved for all scenarios for both systems. For the car-based system, the average TPRs are improved to 65.02$\%$ and 81.65$\%$ in the two scenarios, respectively. Similarly, the average TPRs are improved to 65.46$\%$ and 75.29$\%$ for the wearable system. 

We repeatedly perform a paired two-sample t-test on the TPRs before (baseline) and after the probabilistic fusion process for all the randomly sampled frames from each scenario for both sensing systems (Table \ref{tab:performance_tpr}). All tests reject the null hypothesis with small p-values ($p<0.001$ for the car-based system in both scenarios and the wearable system in scenario 2, and $p<0.01$ for the wearable system in scenario 1). In other words, there is strong evidence to conclude that the differences between the TPRs calculated before and after applying the proposed algorithm are greater than zero. Therefore, the probabilistic fusion process can improve the mapping accuracy significantly. So that hypothesis 1 is proved. 

With the proposed algorithm proved to improve TPRs significantly, it is important to confirm if the improved performance is sufficient for localization purposes. Thus, we conduct a one-sample right-tailed t-test to infer whether the average TPRs after the implementation of the algorithm is greater than 50\%, with results shown in Table \ref{tab:performance_tpr_0.5}. The results show that for both systems, the average TPRs are significantly larger than 50\% ($p-value<0.05$ for scenario 1 and $p-value<0.001$ for scenario 2, respectively). This result proves hypothesis 2. Since the fusion process selects the median value of the mapped LiDAR point for localization, a TPR larger than 50\% can guarantee that a correct LiDAR point is selected, which is reflected by the target object. 

\begin{table}[h]
\centering
\caption{One sample (right-tailed) t-test  results for true positive rate, where $H_1$: $\mu >$0.5. (Mean\_1 refer to the average TPR after the implementation of the algorithm)}
\label{tab:performance_tpr_0.5}
\begin{tabular}{l|lllll}
\hline
System   & \multicolumn{1}{l|}{Sce}  & \multicolumn{1}{l|}{Mean\_1 (\%)} & \multicolumn{1}{l|}{\textit{t}} & \textit{p-val}  & N \\ \hline
Car      & 1                                          & 65.02(19.01)                      & 2.50                            & .017 & 10 \\
         & 2                                          & 81.65(17.93)                      & 7.69                            & \textless{}.001 & 19 \\ \hline
eScooter & 1                                          & 65.46(18.14)                      & 2.09                            & .046            & 6  \\
         & 2                                         & 75.29(12.93)                      & 6.19                           & \textless{}.001 & 10 \\ \hline
\end{tabular}
\end{table}
\subsection{Localization Estimation Accuracy}

In order to further evaluate the fusion performance, we calculate the localization estimation accuracy by comparing the calculated and ground-truth locations of the target objects, using data collected in the two experiment scenarios from both developed sensing systems. The car's ground-truth locations are achieved from the GPS/IMU data, and the estimated locations are calculated from the wearable sensing system's outputs, applying the proposed fusion process. A similar process is conducted for the e-scooter rider's location. 

A total of 102 frames were randomly sampled from all the collected data (57 for the car-based system's output and 45 for the wearable system's outputs). Then, we calculated the mean absolute error (MAE) (using equations \ref{MAEx} and \ref{MAEy}) between the estimated trajectories and the ground-truth trajectories at all frames in meters. The average MAE value for the car-based system is 1.15 meters with a standard deviation of 0.62 meters. The average MAE value for the wearable system is 1.07 meters with a standard deviation of 0.68 meters. Since the localization errors are comparable to the sizes of the target objects, the output can be accurately used to support trajectory generation and crash estimation. 


There are several contributors to the localization errors after applying the proposed fusion process. One significant part of the errors comes from the size of the target objects. The current process does not differentiate the exact reflection point of the object, which can be away from the GPS receiver. Since the ground-truth distance is calculated using the relative positions of GPS receivers with an accuracy of 2 cm, the distance of the actual LiDAR reflection point from the GPS receiver can contribute to some inherent localization errors. This problem is true for both targets of the car or the e-scooter rider. Similarly, the distance between the LiDAR origin and the GPS origin in both sensing systems also contributes to the errors. These two types of errors can be further reduced. However, since the paper's primary purpose is to introduce the probabilistic fusion process, the additional work is out of the current scope and will not be included. 

Another contributing factor to the localization error is synchronization offsets between the LiDAR (10 fps) and GPS (5 fps) units. The two inputs can have up to 0.1 second differences due to the different sampling rates, which will cause errors of about 0.9 meters at the speed of 20 mph. Additional work to reduce this error is also out of the scope of discussing the probabilistic fusion process and will not be discussed.



\section{Conclusion}
As one core component of the AV sensing system, LiDAR-camera fusion attracts much research attention recently. However, although typical calibration can achieve very high fusion accuracy in a static lab environment, inherent mapping errors are inevitable and expensive to deal with in practice. This limitation becomes critical when the sensing target is micro-mobility vehicles like e-scooters which are small and move fast. In this paper, a low-cost probabilistic solution to the mapping errors in LiDAR-camera fusion is proposed for localization purposes. The multiple-step data process is comprised of RANSAC-based data preprocessing, bounding box enlargement, mode-based clustering, shape similarity comparison using KL-divergence, distance and angle calculations, and trajectory smoother utilizing three order-two polynomial regression models with RANSAC.

Two LiDAR-camera sensing systems were developed for cars and wearing, which are carefully calibrated following the typical processes to achieve very high fusion accuracy in the lab environment. Experiments were conducted between car and e-scooter riders with these sensing systems to collect their interaction data in two scenarios. Results have shown that although large mapping errors are observed in the raw data, the proposed fusion process can significantly reduce the mapping errors measured by the TPRs, which can be accurately used for localization, trajectory generation, and crash estimation purposes. 

After applying the proposed method, there are still localization errors due to different coordinate origins between the LiDAR and GPS and their synchronization offsets, which will be further reduced in future work. Moreover, the proposed method will be implemented in real-time in future applications to improve the accuracy of AV perception systems. 


\section*{Acknowledgment}
The authors would like to thank the project sponsor: Collaborative Safety Research Center (CSRC), Toyota Motor North America, Research and Development, Ann Arbor, Michigan.

\ifCLASSOPTIONcaptionsoff
  \newpage
\fi

\vfill


\end{document}